\documentclass[conference]{IEEEtran}
\IEEEoverridecommandlockouts
\usepackage[utf8]{inputenc}
\usepackage{cite}
\usepackage{amsmath,amssymb,amsfonts}
\usepackage{algorithmic}
\usepackage{graphicx}
\usepackage{relsize}
\usepackage{caption}
\usepackage{textcomp}
\usepackage{xcolor}
\usepackage{dblfloatfix}

\def\BibTeX{{\rm B\kern-.05em{\sc i\kern-.025em b}\kern-.08em
    T\kern-.1667em\lower.7ex\hbox{E}\kern-.125emX}}
\begin{document}

\title{On Filter Size in Graph Convolutional Networks
}
\author{
    \IEEEauthorblockN{Dinh V. Tran\IEEEauthorrefmark{1}\IEEEauthorrefmark{2}, Nicolò Navarin\IEEEauthorrefmark{1}\IEEEauthorrefmark{3}, Alessandro Sperduti\IEEEauthorrefmark{1}}
    \IEEEauthorblockA{\IEEEauthorrefmark{1}Department of Mathematics,
University of Padova, Italy \\
\{dinh, nnavarin, sperduti\}@math.unipd.it}
\IEEEauthorblockA{\IEEEauthorrefmark{2} Bioinformatics Group, Department of Computer Science, University of Freiburg, Germany \\
dinh@informatik.uni-freiburg.de}
\IEEEauthorblockA{\IEEEauthorrefmark{3}School of Computer Science, 
University of Nottingham, United Kingdom \\
nicolo.navarin@nottingham.ac.uk}
}


\maketitle

\begin{abstract}
Recently, many researchers have been focusing on the definition of neural networks for graphs. The basic component for many of these approaches remains the graph convolution idea proposed almost a decade ago.
In this paper, we extend this basic component, following an intuition derived from the well-known convolutional filters over multi-dimensional tensors.
In particular, we derive a simple, efficient and effective way to introduce a hyper-parameter on graph convolutions that influences the filter size, i.e. its receptive field over the considered graph.
We show with experimental results on real-world graph datasets that the proposed graph convolutional filter improves the predictive performance of Deep Graph Convolutional Networks.
\end{abstract}

\begin{IEEEkeywords}
graphs, deep learning for graphs, graph convolution, convolutional neural networks for graphs.
\end{IEEEkeywords}

\section{Introduction}
Graphs are a common and natural way to represent many real world data, e.g. in Chemistry a compound can be represented by its molecular graph, in social networks the relationships between users are represented as edges in a graph where users are nodes. Many computational tasks involving such graphical representations require machine learning, such as classification of active/non-active drugs or prediction of the creation of a future link between two users in a social network. State-of-the-art machine learning techniques for classification and regression on graphs are at the moment kernel machines equipped with specifically designed kernels for graphs (e.g,
~\cite{shervashidze2009efficient,vishwanathan2010graph,MartinoNS12}). Although there are examples of kernels for structures that can be designed on the basis of a training set \cite{Fisher, sombased, GenerativeKerneles}, most of the more efficient and effective graph kernels are based on predefined structural features, i.e, features definition is not part of the learning process.  

There is a recent shift of trend from kernels to neural networks for graphs. Unlike kernels, the definition of features in neural networks are defined based on a learning process which is supervised by the graph's labels (targets).
Many approaches have addressed the problem of defining neural networks for graphs~\cite{surveyNN4G}. However, one of the core components, the graph convolution, has not changed much with respect to the earlier works~\cite{Micheli2009,Scarselli2009}.

In this paper, we work on the re-design of this basic component. We propose a new formulation for the graph convolution operator that is strictly more general than the existing one.
Our proposal can be virtually applied to all the techniques based on graph convolutions.

The paper is organized as follows. We start in Section~\ref{sec:def} with some basic definitions and notation. In Section~\ref{sec:graphconvolution}, we provide an overview over the various proposals of graph convolution available in literature. 
In Section~\ref{sec:pargraphconv} we detail our proposed parametric graph convolutional filter.
In Section~\ref{sec:related} we discuss other related works that are not based on graph convolution, including some alternative graph neural network architectures and graph kernels.
In Section~\ref{sec:exps} we report our experimental results. Finally, Section~\ref{sec:conclusions} concludes the paper.
\section{Notation  and Definitions}
\label{sec:def}
We denote matrices with bold uppercase letters, vectors with uppercase letters, and variables with lowercase letters. Given a matrix $\mathbf{M}$, $M_i$ denotes the $i$-th row of the matrix, and $m_{ij}$ is the element in $i$-th row and $j$-th column. Given the vector $V$, $v_i$ refers its $i$-th element. 

Let's consider $G=(V^G, E^G, \mathbf{X}^G)$ as a graph, where \mbox{$V^G=\{v_1, \ldots, v_n\}$} is the set of vertices (or nodes), \mbox{$E_G \subseteq V^G \times V^G$} is the set of edges, and $\mathbf{X}^G \in \mathbb{R}^{n\times d}$ is a node label matrix, where each row is the label (a vector of size $d$) associated to each vertex $v_i \in V^G$ , i.e. \mbox{$X^G_i = (x_{i,0}, \ldots, x_{i,d})$}. Note that, in this paper, we will not consider edge labels.
When the reference to the graph $G$ is clear from the context, for the sake of notation we discard the superscript referring to the specific graph. 
We define the adjacency matrix $A \in \mathbb{R}^{n \times n}$ as ${{a}}_{ij}=1 \iff (i,j) \in E$, 0 otherwise.
We also define the \emph{neighborhood} of a vertex $v$ as the set of vertices connected to $v$ by an edge, i.e. $\mathcal{N}(v)=\{u | (v,u) \in E\}$. Note that $N(v)$ is also the set of nodes at shortest path distance exactly one from $v$, i.e. $\mathcal{N}(v)=\{u | sp(v,u)=1\}$, where $sp$ is a function computing the shortest-path distance between two nodes in a graph.

In this paper, we consider the problem of graph classification. Given a dataset composed of $N$ pairs \mbox{$\{(G_i, y_i) | 1 \leq i \leq N\}$}, the task is then, given an unseen graph G, to predict its correct target $y$.

\section{Graph convolutions}
\label{sec:graphconvolution}
 The first 
 definition of neural network for graphs has been proposed in~\cite{SperdutiStarita}. More recent models have been 
proposed in \cite{Micheli2009,Scarselli2009}.
Both works are based on an idea that has been re-branded later as \emph{graph convolution}.

The idea is to define the neural architecture following the topology of the graph. Then a transformation is performed from the neurons corresponding to a vertex and its neighborhood to a hidden representation, that is associated to the same vertex (possibly in another layer of the network).
This transformation depends on some parameters, that are shared among all the nodes.
In the following, for the sake of simplicity we ignore the bias terms. 

In \cite{Scarselli2009}, when considering non-positional graphs, i.e. the most common definition, and the one we are considering in this paper, a transition function on a graph node $v$ at time $0\leq t$ is defined as:
\begin{equation}
H^{t+1}_v =\sum_{u \in \mathcal{N}(v)} f_{\Theta}({H}^{t}_u,{X}_v,{X}_u),
\label{eq:scarsellirec}
\end{equation}
where $f_\Theta$ is a parametric function whose parameters $\Theta$ have to be learned (e.g. a neural network) and are shared among all the vertices.
Note that, if edge labels are available, they can be included in eq.~\eqref{eq:scarsellirec}. In fact, in the original formulation, $f_{\Theta}$ depends also on the label of the edge between $v$ and $u$.
This transition function is part of a recurrent system. It is defined as a contraction mapping, thus the system is guaranteed to converge to a fixed point, i.e. a representation, that does not depend on the particular initialization of the weight matrix $\mathbf{H}^0$.
The output is computed from the last representation and the original node labels as follows:
\begin{equation}
O^t_v=g_{\Theta'}(H^t_v,X_v),
\end{equation}
where $g_{\Theta'}$ is another neural network.
\cite{Li2015b} extends the work in \cite{Scarselli2009} by removing the constraint for the recurrent system to be a contraction mapping, and replacing the recurrent units with GRUs.
However, recently it has been shown in~\cite{Bresson2018} that stacked graph convolutions are superior to graph recurrent architectures in terms of both accuracy and computational cost.

In \cite{Micheli2009}, a model referred to as Neural Network for Graphs (NN4G) is proposed. In the first layer, a transformation over node labels is computed:
\begin{equation}
\hat{h}^1_v=f \left ( \sum_{j=1}^{d} \bar{w}_{1,j} x_{v,j} \right ),
\label{eq:micheliconv1}
\end{equation}
where $\bar{W}_1$ are the weights connecting the original labels $X$ to the current neuron, and $1 \leq v \leq n$ is the vertex index.
The graph convolution is then defined for the $i+1$-th layer (for $i>0$) as:
\begin{equation}
\hat{h}^{i+1}_v=f \left ( \sum_{j=1}^{d} \bar{w}_{i+1,j} x_{v,j}+
\sum_{k=1}^{i} \hat{w}_{i+1,k}\sum_{u \in \mathcal{N}(v)} \hat{h}^k_u
\right ),
\label{eq:micheliconv}
\end{equation}
where $\hat{W}_{i+1}$ are weights connecting the previous hidden layers to the current neuron (shared).
Note that in this formulation, skip connections are present, to the $(i+1)$-th layer, from layer $1$ to layer $i$. There is an interesting recent work about the parallel between skip-connections (residual networks in that case) and recurrent networks~\cite{Liao2016}. However, since in the formulation in eq.~\eqref{eq:micheliconv}, every layer is connected to all the subsequent layers, it is not possible to reconduct it to a (vanilla) recurrent model.
Let us consider the \mbox{$(i+1)$-th} graph convolutional layer, that comprehends $c_{i+1}$ graph convolutional filters. We can rewrite eq.~\eqref{eq:micheliconv} for the whole layer as:
\begin{equation}
\mathbf{H}^{i+1}= f(\mathbf{X} \mathbf{\bar{W}}^{i+1} + \sum_{k=1}^{i}\mathbf{A} \mathbf{H}^k \mathbf{\hat{W}}^{i+1,k}),
\label{eq:micheliconvmatrix}
\end{equation}
where ${i={0,\ldots,l-1}}$ (and $l$ is the number of layers), \mbox{$\mathbf{\bar{W}}^{i+1} \in \mathbb{R}^{d \times c_{i+1}}$}, \mbox{$\mathbf{\hat{W}}^{{i+1},k} \in \mathbb{R}^{c_{k} \times c_{i+1}}$}, $\mathbf{H}^k \in \mathbb{R}^{n \times c_k}$, $c_i$ is the size of the hidden representation at the $i$-th layer, and $f$ is applied element-wise.
\begin{figure}
\centering
\includegraphics[width=.8\linewidth]{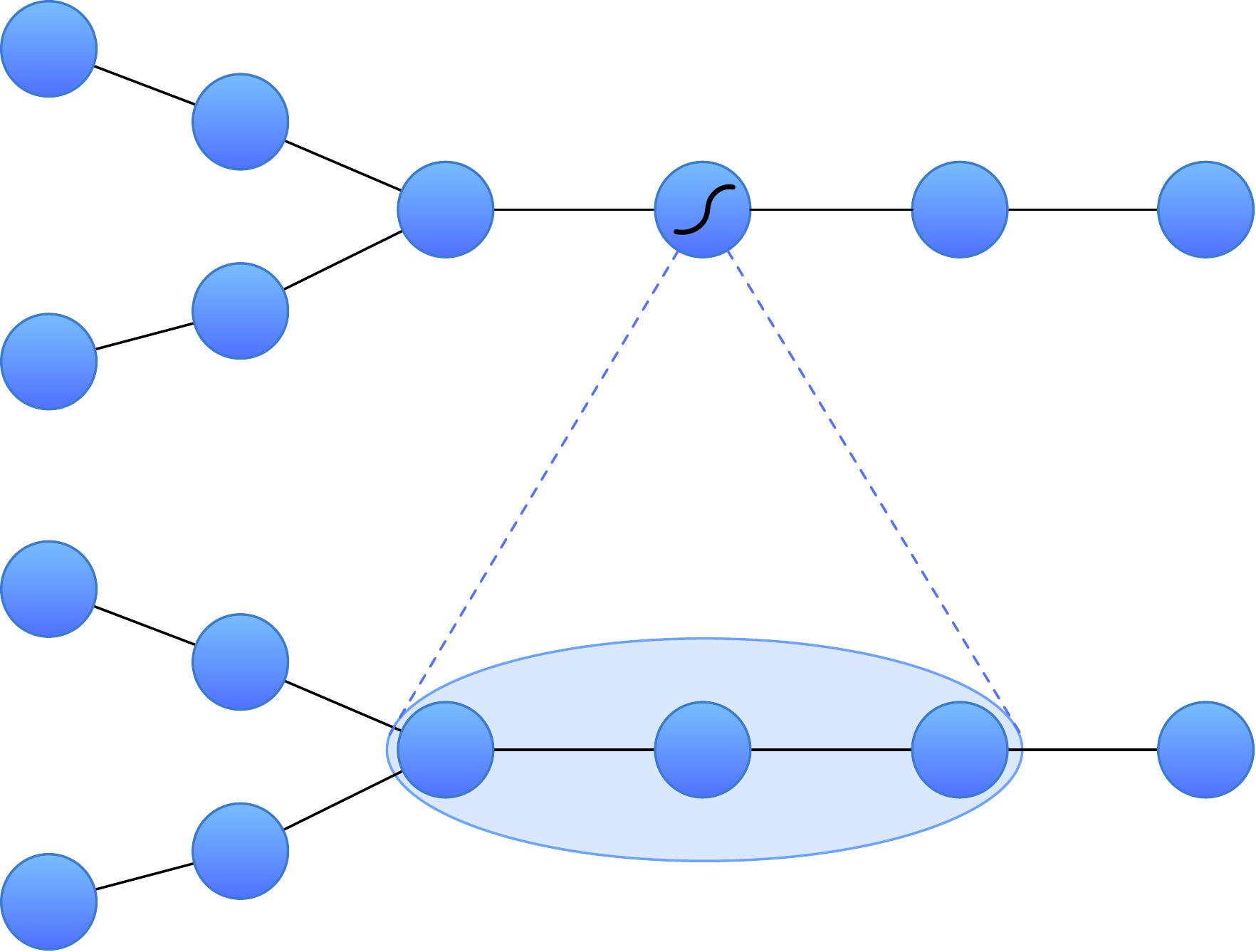}
\caption{Graph convolution as described in \cite{Micheli2009}, and adopted with some variations by many state-of-the-art Graph Convolutional neural networks. \label{fig:conv}}
\end{figure}

An abstract representation of eq.~\eqref{eq:micheliconv} is depicted in Figure~\ref{fig:conv}.
The convolution in eq.~\eqref{eq:micheliconv} is part of a multi-layer architecture, where each layer's connectivity resembles the topology of the graph, and the training is layer-wise. Finally, for each graph, NN4G computes the average graph node representation for each hidden layer, and concatenates them. This is the graph representation computed by \textit{NN4G}, and it can be used for the final prediction of graph properties with a standard output layer.

In \cite{Duvenaud2015b}, a hierarchical approach has been proposed. This method is similar to NN4G and is inspired by circular fingerprints in chemical structures. While \cite{Micheli2009} adopts Cascade-Correlation for training, \cite{Duvenaud2015b} uses an end-to-end back-propagation.
 ECC \cite{simonovsky2017dynamic} proposes an improvement of \cite{Duvenaud2015b}, weighting the sum over the neighbors of a node by weights conditioned by the edge labels. We consider this last version as a baseline in our experiments.

Recently, \cite{Kipf2016a} derives a graph convolution that closely resembles~\eqref{eq:micheliconv}. Let us, from now on, consider $\mathbf{H}^0=\mathbf{X}$.
Motivated by a first-order approximation of localized spectral filters on graphs, the proposed graph convolutional filter looks like:
\begin{equation}
\mathbf{H}^{i+1}= f(\mathbf{\tilde{D}}^{-\frac{1}{2}} \mathbf{\tilde{A}} \mathbf{\tilde{D}}^{-\frac{1}{2}} \mathbf{H}^i \mathbf{W}^i),
\label{eq:kipf}
\end{equation}
where $\mathbf{\tilde{A}}=\mathbf{A}+\mathbf{I}$, $\tilde{d}_{ii} = \sum_j \tilde{{a}}_{i,j}$, and $f$ is any activation function applied element-wise.

If we ignore the terms $\mathbf{\tilde{D}}^{-\frac{1}{2}}$ (that in practice act as normalization), it is easy to see that eq.~\eqref{eq:kipf} is very similar to  eq.~\eqref{eq:micheliconvmatrix}, the difference being that there are no skip connections in this case, i.e. the $(i+1)$-th layer is connected just to the $i$-th layer. Consequently, we just have to learn one weight matrix per layer.

In \cite{Zhang2018}, a slightly more complex model compared to \cite{Kipf2016a} is proposed. This model shows the highest predictive performance with respect to the other methods presented in this section. The first layers of the network are again stacked graph convolutional layers, defined as follows:
\begin{equation}
\textbf{H}^{i+1}=f(\mathbf{\tilde{D}}^{-1}\mathbf{\tilde{A}} \mathbf{H}^{i} \mathbf{W}^{i}),
\label{eq:dgcnn}
\end{equation}
where $\mathbf{H}^0=\mathbf{X}$ and $\mathbf{\tilde{A}}=\mathbf{A}+\mathbf{I}$. Note that in the previous equation, we compute the representation of all the nodes in the graph at once.
The difference between eq.~\eqref{eq:dgcnn} and eq.~\eqref{eq:kipf} is the use of different propagation scheme for nodes' representations: eq.~\eqref{eq:kipf} is based on the normalized graph Laplacian, while eq.~\eqref{eq:dgcnn} is based on the random-walk graph Laplacian. In \cite{Zhang2018}, authors state that the choice of normalization does not significantly affect the results.
In fact, both equations can be seen as  first-order approximations of the polynomially parameterized spectral graph convolution.
In \cite{Zhang2018}, three graph convolutional layers are stacked.
The graph convolutions are followed by a concatenation layer that merges the representations computed by each graph convolutional layer. Then, differently from previous approaches, the paper introduces a \emph{sortpooling} layer, that selects a fixed number of node representations, and computes the output from them stacking 1D convolutional layers and dense layers.
This is the same network architecture that we considered in this paper.

\subsection{SortPooling layer}
After stacking some graph convolution layer, we need a mechanism to predict the target for the graph, starting from its node encoding. Ideally, this mechanism should be applicable to graphs with variable number of vertices.
Instead of averaging the node representations, \cite{Zhang2018} proposes to solve this issue with the \textit{SortPooling} layer.

Let us assume that the encoding, for each node, of the $i$-th graph convolution layer is $c$.
Let us  consider the output of the last graph convolution (or concatenation) layer to be \mbox{$\mathbf{H}^l \in \mathbb{R}^{n \times c}$}
, where
each row is a vertex's feature descriptor and each column is a feature channel. The output of the SortPooling layer is a $k \times c$
tensor, where $k$ is a user-defined integer.

In the SortPooling layer, the rows of $\mathbf{H}^l$ 
are sorted lexicographically (possibly starting from the last column). We can see the output of the graph convolutional layer as continuous WL colors, and thus we are sorting all the vertices according to these colors. This way, a consistent ordering is imposed for graph vertices, making it possible to train traditional neural networks on the sorted graph representations.

In addition to sorting vertex features in a consistent order,
the other function of SortPooling is to unify the sizes of the output tensors. After sorting, we truncate/extend the output tensor in the first dimension from n to k. The intention is to unify graph sizes, making graphs with different numbers of vertices unify their sizes to k. The unifying is done by
deleting the last $n-k$ rows if $n > k$, or adding $k-n$ zero rows if $n < k$.

Note that if two vertices have the same hidden representation, it doesn't matter which node we pick because the output of the SortPooling layer would be exactly the same.

\section{Parametric Graph Convolutions}
\label{sec:pargraphconv}

{A straightforward generalization of eq.~\eqref{eq:dgcnn} would be defined on the powers of the adjacency matrix, i.e. on random walks~\cite{Abu-El-Haija2018}. This would introduce tottering in the learned representation, that is not considered to be beneficial in general. We decided to follow another approach, based on shortest-paths.}
As mentioned before, the adjacency matrix $\mathbf{A}$ of a graph can be seen as the matrix of the shortest-paths of length $1$, i.e.
\begin{equation}
\mathbf{a}_{i,j}=sp^1_{i,j} =\begin{cases}
1 & \text{if } sp(i,j)=1\\
0 & \text{otherwise}\\
\end{cases}.
\end{equation}
Moreover, the identity matrix $\textbf{I}$ is the matrix of the shortest-paths of length $0$ (assuming that each node is at distance zero from itself), i.e. $\textbf{I}=\mathbf{SP}^0$. Moreover, note that \mbox{$\mathbf{\tilde{A}}= \mathbf{SP}^0 + \mathbf{SP}^1$}.

By means of this new notation, we can rewrite eq.~\eqref{eq:dgcnn} as:
\begin{equation}
\textbf{H}^{l+1}=f \left (\mathbf{\tilde{D}}^{-1} \left (\mathbf{SP}^0 +\mathbf{SP}^1  \right ) \mathbf{H}^{l} \mathbf{W}^{l} \right ).
\label{eq:ourdgcnn}
\end{equation}

Let us now define ${\hat{d}}^r_{ii} = \sum_j {sp}^r_{i,j}$.
We can now extend our reasoning and define our \textit{parameterized} (by $r$) graph convolution layer.
{
In our contribution, we decided to process information in a slightly different way with respect to~\eqref{eq:ourdgcnn}. Instead of summing the contributions of the $\mathbf{SP}$ matrices, we decided to keep the contributions of the nodes at different shortest-path distance separated. This is equivalent to the definition of multiple graph convolutional filters, one for each shortest-path distance. We define the Parametric Graph Convolution as:}

\begin{equation}
\mathbf{H}^{r,l+1} = \mathlarger{ \|}_{j=0}^r f \left ( (\mathbf{\hat{D}}^j)^{-1} \mathbf{SP}^j \mathbf{H}^{l} \mathbf{W}^{j,l} \right ),
\label{eq:our}
\end{equation}
where $\|$ is the vertical concatenation of vectors.
Note that with our formulation, we have a different $\mathbf{W}^{j,l}$ matrix for each layer $l$ and for each shortest-path distance $j$. Moreover, as mentioned before, we are concatenating the information and not summing it, explicitly keeping the contributions of the different distances separated. This approach follows the network-in-network idea~\cite{Szegedy2015a}. In our case, at each layer, we are effectively applying, at the same time, $r+1$ convolutions (one for each shortest-path distance) and concatenating their output.
Let us fix a parameter controlling the number of filters for the $l$ layer, say $c_l$, and a value for the hyper-parameter $r$, then we have $\mathbf{H}^{r,l+1} \in \mathbb{R}^{n \times r\cdot c_l}$.

\begin{figure}
\centering
\includegraphics[width=.8\linewidth]{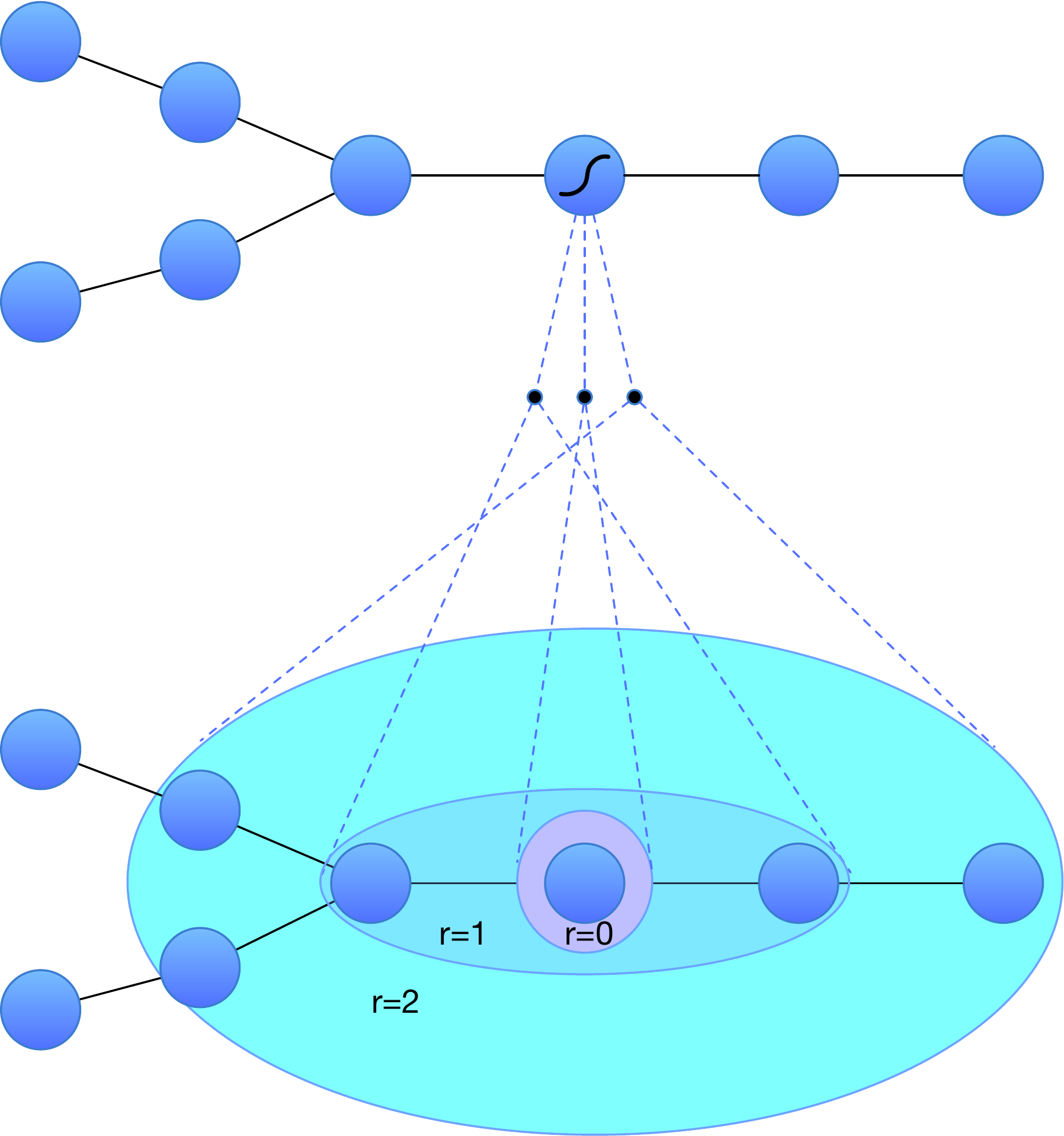}
\caption{The proposed Parametric Graph Convolution. The parameter $r$ controls the maximum distance of the considered neighborhood, and the dimensionality of the output. \label{fig:our}}
\end{figure}
\subsection{Receptive field}
It has been shown in \cite{Kipf2016a,Zhang2018} that with the standard definition of graph convolution, e.g. the ones in eq.~\eqref{eq:kipf} and eq.~\eqref{eq:dgcnn}, the receptive field of a graph convolutional filter at layer $l$ corresponding to the vertex $v$ is $\mathcal{N}_l(v)$. This draws an interesting parallel with the Weisfeiler-Lehman graph kernel {(see Section~\ref{sec:gk})}, where intuitively the number of WL iterations is equivalent to the number of stacked graph convolution layers in the architecture.

In our proposed parametric graph convolution in eq.~\eqref{eq:our}, the $r$ parameter directly influences the considered neighborhood in the graph convolutional filter (and the number of output channels, since we concatenate the output of the convolutions 
{for all $j \leq r$}). It is easy to see that, by definition, the receptive field of a  graph convolutional filter parameterized by $r$ and applied to the vertex $v$ includes all the nodes at shortest-path distance at most $r$ from $v$.
When we stack multiple layers of our parametric graph convolution, the receptive field grows in the same way. 
The receptive field of a parametric graph convolutional filter of size $r$ at layer $l$ applied to the vertex $v$ includes then all the vertices at shortest-path distance at most $l \cdot r$ from $v$.


\subsection{Computational complexity}
Equation~\eqref{eq:our} requires to compute the all-pairs shortest paths, up to a fixed length $r$. 
While computing the unbounded shortest paths for a graph with $n$ nodes requires $O(n^3)$ time, if the maximum length is small enough, it is possible to implement it with one depth-limited breadth-first visit 
starting from each node, with an overall complexity of $O(m^r)$, where $m$ is the number of edges in a graph.

\section{Related works}
\label{sec:related}
Besides the approaches based on graph convolutions presented in Section~\ref{sec:graphconvolution}, there are some other 
{methods}
in literature to process graphs with neural networks.

For instance, \cite{Velickovic2017} defined an attention mechanism to propagate information between the nodes in a graph.
The basic idea is the definition of an external network that, given two neighboring nodes, outputs an attention weight for that specific edge. 
A shared attentive mechanism $a : \mathbb{R}^{d} \times \mathbb{R}^{d} \rightarrow \mathbb{R}$
computes the attention coefficients
\begin{equation}
e_{v,u} = a_{\Theta}(\mathbf{W} X_v,\mathbf{W} X_u),
\end{equation}
that indicate the importance of node $u$’s features to node $v$.
Here, $a_{\Theta}$ is a parametric function, that in the original paper is a single-layer feed-forward network parameterized by the vector $\Theta \in \mathbb{R}^{2d}$.
The information about the graph structure is injected into the mechanism by performing masked attention, i.e. $\mathbf{e}_{v,u}$ is only computed for nodes
$u \in \mathcal{N}(v)$. To make coefficients easily comparable across different nodes, a softmax function is used:
\begin{equation}
b_{v,u} = softmax_u (e_{v,u}) = \frac{exp({e}_{v,u})}{ \sum_{k\in \mathcal{N}(v)} exp({e}_{v,k})}.
\end{equation}


Once obtained, the normalized attention coefficients are used to compute a linear combination of the features corresponding to them, to serve as the final output features for every node (after potentially applying a point-wise nonlinearity, $f$):
\begin{equation}
H_v = f
\left( \sum_{u \in \mathcal{N}(v)} b_{vu} \mathbf{W} X_u \right ) .
\label{eq:attention}
\end{equation}

To stabilize the learning process of self-attention, authors propose to extending the mechanism to employ multi-head attention($K$ different attention weights per edge). 
For the last layer, authors employ averaging, and delay applying the final nonlinearity (usually a softmax or logistic sigmoid for classification problems) until then.

This technique has been applied to node classification only, and its complexity (due to implementation issues) is high.
In principle, the same approach of NN4G can be adopted to generate graph-level representations and predictions for this model.

\cite{niepert2016learning} (PSCN) proposes another interpretation of graph convolution. Given a graph, it first selects the nodes where the convolutional filter have to be centered. Then, it selects a fixed number of vertices from its neighborhood, and infers an order on them. This ordering constraint limits the flexibility of the approach because learning a consistent order is difficult, and the number of nodes in the convolutional filter {has} to be fixed a-priori.

 Diffusion CNN (DCNN) \cite{atwood2016diffusion} is based on the principle of heat diffusion (on graphs).  The idea is to map from nodes and their labels to the result
of a diffusion process that begins at that node. 
\begin{table*}[!t]
\def\arraystretch{1.6}
\caption{Summary of employed graph datasets}
\label{tab:datasets}
\begin{center}
\begin{tabular}{lcccccccc}
\hline
\textbf{Dataset} & \textbf{MUTAG} & \textbf{PTC} & \textbf{NCI1} & \textbf{PROTEINS} & \textbf{D$\&$D} & \textbf{COLLAB} & \textbf{IMDB-B} & \textbf{IMDB-M} \\
\hline
$\#$Nodes (Max) & 28  & 109  & 111 & 620 & 5748 & 492 & 136 & 89\\
$\#$Nodes (Avg) & 17.93 & 25.56 & 29.87 & 39.06 & 284.32 & 74.49 & 19.77 & 13.00\\
$\#$Graphs 		& 188   & 344 & 4110 & 1113 & 1178 & 5000 & 1000 & 1500\\
\hline

\end{tabular}
\label{tab1}
\end{center}
\end{table*}
\subsection{Graph Kernels}
\label{sec:gk}
Kernel methods defines the model as a linear classifier in a Reproducing Kernel Hilbert Space, that is the space implicitly defined by a kernel function $\mathcal{K}(x_1,x_2)= \langle \phi(x_1),\phi(x_2)\rangle$. SVM is the most popular kernelized learning algorithm, that defines the solution as the maximum-margin hyper-plane.

Kernel functions can be defined for many objects, and in particular for graphs.
Many graph kernels have been defined in literature. For instance, Random Walk kernels are based on the number of common random walks in two graphs~\cite{Gartner2003a,vishwanathan2010graph}  and can be computed efficiently in closed form.
More recent proposals focus on more complex structures, and allow to represent the $\phi$ function explicitly, with computational benefits. Among others, kernels have been defined considering graphlets~\cite{Shervashidze2009}, shortest-paths~\cite{Kriegel05shortestpath},  subtrees~\cite{DaSanMartino2016, Dasan2012} and subtree-walks~\cite{DaSanMartino2014a,Shervashidze2011}. 
{For instance, the Weisfeiler-Lehman subtree kernel (WL) defines its features as rooted subtree-walks, i.e, subtrees whose nodes can appear multiple times, up to a user-defined maximum height $h$ (maximum number of iterations).}

Propagation kernels (PK)~\cite{neumann2012efficient} follow a different idea, inspired by the diffusion process in graph node kernels (i.e. kernels between nodes in a single graph), of propagating the node label information through the edges in a graph. Then, for each node, a distribution over the propagated labels is computed. Finally, the kernel between two graphs compares such distributions over all the nodes in the two graphs.

While exhibiting state-of-the-art performance on many graph datasets, the main problem of graph kernels is that they define a fixed representation, that is not task-dependent and can in principle limit the predictive performance of the method.
Deep graph kernels (DGK)~\cite{yanardag2015deep} propose an approach to alleviate this problem.
Let us fix a base kernel and its explicit representation $\phi(\cdot)$. Then a deep graph kernel can be defined as:
$$
\mathcal{DGK}(x_1,x_2) =  \phi(x_1)^T \mathbf{{M}} \phi(x_2),
$$
where $\mathbf{{M}}$ is a matrix of parameters that has to be learned, possibly including target information.

\section{Experiments}
\label{sec:exps}
In this section, we aim at evaluating the performance of the proposed method and comparing it with many existing graph kernels and deep learning approaches for graphs. We pay a special attention to the performances of our method and DGCNN, to see whether the proposed generalization helps to improve the predictive performance. As a means to achieve this purpose, various experiments are conducted in two settings, following the experimental procedure used in \cite{Zhang2018} on eight graph datasets (see Table~\ref{tab:datasets} for a summary). The code for our experiments is available online at \textit{https://github.com/dinhinfotech/PGC-DGCNN}.

In the first setting, we compare the performance of our method with DGCNN and state-of-the-art graph kernels: the graphlet kernel (GK) \cite{shervashidze2009efficient}, the random walk kernel (RW) \cite{vishwanathan2010graph}, the propagation kernel (PK) \cite{neumann2012efficient}, and the Weisfeiler-Lehman subtree kernel (WL) \cite{shervashidze2011weisfeiler}.
We do not include other state-of-the-art graph kernels  such as NSPDK~\cite{Costa2010} and ODD \cite{DaSanMartino2016} because their performance is not much different from the considered ones, and it is above the scope of this paper to extensively compare the graph kernels in literature.
In this setting, five datasets containing biological node-labeled graphs are employed, namely MUTAG~\cite{Debnath1991}, PTC~\cite{Toivonen2003}, NCI1~\cite{springerlink:10.1007/s10115-007-0103-5}, PROTEINS, and D$\&$D~\cite{dobson2003}.
In the first three datasets, each graph represents a chemical compound, where nodes are labeled with the atom type, and edges represent bonds between them.
MUTAG is a dataset of aromatic and hetero-aromatic nitro compounds, where the task is to predict their mutagenic effect on a bacterium.
In PTC, the task is to predict chemical compounds carcinogenicity for male and female rats. NCI1 contains anti-cancer screens for cell lung cancer.
In PROTEINS and D$\&$D, each graph represents a protein. The nodes are labeled according to the amino-acid type. The proteins are classified into two classes: enzymes and non-enzymes.

In the second setting, we desire to evaluate the performance of the proposed method and DGCNN along with other deep learning approaches for graphs: PATCHY-SAN (PSCN) \cite{niepert2016learning}, Diffusion CNN (DCNN) \cite{atwood2016diffusion}, ECC \cite{simonovsky2017dynamic} and Deep Graphlet Kernel (DGK) \cite{yanardag2015deep}. In this setting, three biological datasets (NCI1, PROTEINS and D$\&$D) and three social network datasets from~\cite{yanardag2015deep} (COLLAB, IMDB-B and IMDB-M) are used. 
COLLAB is a dataset of scientific collaborations, where ego-networks are generated for researchers and are classified in three research fields. IMDB-B (binary) is a movie collaboration dataset where ego-networks for actors/actresses are classified in \emph{action} or \emph{romance} genres. IMDB-M is a multi-class version of IMDB-B, containing genres \emph{comedy}, \emph{romance}, and \emph{sci-fi}.
 
In this setting, we eliminate MUTAG and PTC since they have a small number of examples which easily causes over-fitting problems for deep learning approaches.
\begin{table*}[t]
\def\arraystretch{1.6}
\begin{center}
\captionsetup{width=.75\textwidth}
\caption{Comparison with graph kernels. $^*$: our proposed approach. DGCNN is similar to our approach with $r=1$.}
\label{tab:reuslts-kernels}
\begin{tabular}{lccccc}
\hline
\textbf{Dataset} & \textbf{MUTAG} & \textbf{PTC} & \textbf{NCI1} & \textbf{PROTEINS} & \textbf{D$\&$D}  \\
\hline
GK    & 81.39$\pm$1.74 & 55.65$\pm$0.46 & 62.49$\pm$0.27 & 71.39$\pm$0.31 & 74.38$\pm$0.69 \\ 
RW    & 79.17$\pm$2.07 & 55.91$\pm$0.32 & $>$3 days        & 59.57$\pm$0.09 & $>$3 days    \\
PK    & 76.00$\pm$2.69 & 59.50$\pm$2.44 & 82.54$\pm$0.47 & 73.68$\pm$0.68 & 78.25$\pm$0.51 \\
WL    & 84.11$\pm$1.91 & 57.97$\pm$2.49 & \textbf{84.46$\pm$0.45} & 74.68$\pm$0.49 & 78.34$\pm$0.62 \\
\hline
DGCNN & 85.83$\pm$1.66 & 58.59$\pm$2.47 & 74.44$\pm$0.47 & 75.54$\pm$0.94 & \textbf{79.37$\pm$0.94} \\
PGC-DGCNN$^*$ ($r=2$)   & \textbf{87.22$\pm$1.43} & \textbf{61.06$\pm$1.83} & 76.13$\pm$0.73 & \textbf{76.45$\pm$1.02} & 78.93$\pm$0.91 \\
\hline

\end{tabular}
\end{center}
\end{table*}
\begin{table*}[t]
\def\arraystretch{1.6}
\captionsetup{width=.75\textwidth}
\caption{Comparison with other deep learning approaches. $^*$: our proposed approach. DGCNN is similar to our approach with $r=1$.}
\label{tab:results-dnn}
\begin{center}
\begin{tabular}{lcccccc}
\hline
\textbf{Dataset} & \textbf{NCI1} & \textbf{PROTEINS} & \textbf{D$\&$D} & \textbf{COLLAB} &  \textbf{IMDB-B} & \textbf{IMDB-M}\\
\hline
PSCN  & \textbf{76.34$\pm$1.68} & 75.00$\pm$2.51 & 76.27$\pm$2.64 & 72.60$\pm$2.15 & 71.00$\pm$2.29 & 45.23$\pm$2.84 \\
DCNN  & 56.61$\pm$1.04 & 61.29$\pm$1.60 & 58.09$\pm$0.53 & 52.11$\pm$0.71 & 49.06$\pm$1.37 & 33.49$\pm$1.42 \\
ECC   & 76.82          & --              & 72.54          & --              & --              & -- \\
DGK   & 62.48$\pm$0.25 & 71.68$\pm$0.50 & --              & 73.09$\pm$0.25 & 66.96$\pm$0.56 & 44.55$\pm$0.52 \\
\hline
DGCNN & 74.44$\pm$0.47 & 75.54$\pm$0.94 & \textbf{79.37$\pm$0.94} & 73.76$\pm$0.49 & 70.03$\pm$0.86 & \textbf{47.83$\pm$0.85} \\
PGC-DGCNN$^*$   & 76.13$\pm$0.73 & \textbf{76.45$\pm$1.02} & 78.93$\pm$0.91 & \textbf{75.00$\pm$0.58} & \textbf{71.62$\pm$1.22} & 47.25$\pm$1.44\\
\hline
\end{tabular}
\end{center}
\end{table*}
\subsubsection*{Evaluation method and model selection} to evaluate the different methods, a nested 10-fold cross-validation is employed, i.e, one fold for testing, 9 folds for training of which one is used as validation set for model selection. For each dataset, we repeated each experiment 10 times and report the average accuracy over the 100 resulting folds. To select the best model, the hyper-parameters' values of different kernels are set as follows: the height of WL and PK in $\left\lbrace 0, 1, 2, 3, 4, 5 \right\rbrace$, the bin width of PK to 0.001, the size of the graphlets in GK to 3 and the  decay of RW to the largest power of 10 that is smaller than the reciprocal of the squared maximum node degree. Note that some of our results are reported from~\cite{Zhang2018}.
\subsubsection*{Network architecture} we employ the network architecture used in \cite{Zhang2018} to have a fair comparison with DGCNN. The network consists of three graph convolution layers, a concatenation layer, a SortPooling layer, followed by two 1-D convolutional layers and one dense layer. 
The activation function for the graph convolutions is the \emph{hyperbolic tangent}, while the 1D convolutions and the dense layer use \emph{rectified linear units}. Note that our proposal, as presented in Section \ref{sec:pargraphconv}, is a generalization of DGCNN. In other words, DGCNN is very similar to a special case of our method where just the neighbors at shortest-path distance $1$ are considered.
On the contrary, our proposal considers the distance as a hyper-parameter, $r$, allowing to flexibly capture local structures associated to graph nodes. In this section, we set $r$ equal to 2 as the first attempt and plan to explore neighborhoods with nodes at a higher distance as a future work.

\subsection{Experimental Results}
Table \ref{tab:reuslts-kernels} and \ref{tab:results-dnn} show the performance of various methods in the first and second settings, respectively. Overall, DGCNN and our proposed method outperform the compared kernels and deep learning methods in most datasets. 

As can be seen from Table~\ref{tab:reuslts-kernels}, DGCNN and the proposed method (PGC-DGCNN) present higher performances in four out of five datasets with an improvement ranging from 1.03$\%$ to 3.11$\%$ with respect to the best performing kernel. Compared to the RW kernel, our proposed method impressively achieves the highest improvement in MUTAG and PROTEINS with about 8$\%$ and 17$\%$, respectively. Concerning PK and WL, that are similar in spirit to DGCNN and our method as shown in \cite{Zhang2018}, DGCNN and PGC-DGCNN 
illustrate higher performances in most cases with a bigger difference comparing with PK. It is worth noticing, when comparing with PK and WL, that their optimal models in each experiment are selected by tuning the height parameter, $h$, from a range of pre-defined values. { Instead, DGCNN and our method are evaluated with a fixed number of layers only. This indicates that the performance of DGCNN and the proposed method can be higher if we validate the number of stacked graph convolutional layers.}

Related to the performances of various deep learning methods described in Table~\ref{tab:results-dnn}, our method and DGCNN obtain the highest results in five out of six cases, except in NCI1 where they show marginally lower results. Considering the performance of DCNN, DGCNN and our method gain dramatically higher accuracies with the improvement ranges from around 14$\%$ and up to 21$\%$.

We now move our consideration to the difference between the performances of DGCNN and our proposal. It can be seen from the Table~\ref{tab:reuslts-kernels} and \ref{tab:results-dnn} that our method performs better than  DGCNN in the majority of the datasets. In particular, PGC-DGCNN outperforms DGCNN in six out of eight cases with a consistent improvement from about 1$\%$ to 2$\%$. In D$\&$D and IMDB-M, the accuracy of our method is slightly lower than DGCNN. However, these declines are only marginal. The general improved performance of our method comparing to DGCNN can be explained by the fact that our method parameterizes the graph convolutions, making it a generalization of DGCNN. (we recall that we fix the neighborhood distance $r= 2$). In this case, our method captures more information about the local graph structure associated to each node comparing to DGCNN which considers just the direct neighbors, i.e.
$r=1$. It is worth to notice 
that (1) we use a single value for $r$ to build our model. However, in general, we can choose an optimal model by tuning values from a range of values for $r$; (2) we utilize the architecture as proposed in \cite{Zhang2018}, meaning that we have not tried to optimize the network architecture. Therefore, the performance of our method can improve if we optimize the distance parameter $r$ and the number of graph convolutional layers, together with the rest of the architecture.

\section{Conclusions and Future Works}
\label{sec:conclusions}
In this paper, we presented a new definition of graph convolutional filter. It generalizes the most commonly adopted filter, adding an hyper-parameter controlling the distance of the considered neighborhood.
Experimental results show that our proposed filter improves the predictive performance of Deep graph Convolutional Neural Networks on many real-world datasets.

In future, we plan to analyze more in depth the impact of filter size in graph convolutional networks. We will define the 1D convolutions as special cases of Graph Convolutions, and we will explore Fully Graph-Convolutional neural architectures, that will avoid fully-connected layers, and possibly stack more graph convolution layers. Moreover, we will explore the impact of different activation functions for the graph convolutions in such a setting. Finally, we plan to enhance the input graph representation associating to each node the explicit features extracted by graph kernels.

\section*{Acknowledgment}
This project was funded, in part, by the Department of Mathematics, University of Padova, under the DEEP project and DFG project, BA 2168/3-3.

\bibliographystyle{IEEEtran}
\bibliography{bibliography}
\end{document}